\documentclass{article}
\usepackage{spconf,amsmath,graphicx}
\usepackage{multirow}
\usepackage{graphicx}
\usepackage{url}

\title{Inappropriate Pause Detection in Dysarthric Speech Using Large-Scale Speech Recognition}
\name{Jeehyun Lee$^{1*}$ \  Yerin Choi$^{1}$\sthanks{Equal Contribution}\  Tae-Jin Song$^2$ \  Myoung-Wan Koo$^1$\sthanks{This work was supported by Institute of Information \& communications Technology Planning \& Evaluation (IITP) grant funded by the Korea government(MSIT) (No.2022-0-00621, Development of artificial intelligence technology that provides dialog-based multi-modal explainability) }}
\address{Department of Artificial Intelligence, Sogang University, Seoul, South Korea$^1$\\
Department of Neurology, Seoul Hospital, \\Ewha Womans University College of Medicine, Seoul, Republic of Korea$^2$}
\begin{document}
%

\maketitle

\begin{abstract} 
Dysarthria, a common issue among stroke patients, severely impacts speech intelligibility. Inappropriate pauses are crucial indicators in severity assessment and speech-language therapy. We propose to extend a large-scale speech recognition model for inappropriate pause detection in dysarthric speech. To this end, we propose task design, labeling strategy, and a speech recognition model with an inappropriate pause prediction layer. First, we treat pause detection as speech recognition, using an automatic speech recognition (ASR) model to convert speech into text with pause tags. According to the newly designed task, we label pause locations at the text level and their appropriateness. We collaborate with speech-language pathologists to establish labeling criteria, ensuring high-quality annotated data. Finally, we extend the ASR model with an inappropriate pause prediction layer for end-to-end inappropriate pause detection. Moreover, we propose a task-tailored metric for evaluating inappropriate pause detection independent of ASR performance. Our experiments show that the proposed method better detects inappropriate pauses in dysarthric speech than baselines. (Inappropriate Pause Error Rate: 14.47\%)

\end{abstract}

\section{Introduction}
\label{sec:intro}

Post-stroke dysarthria, a speech disorder from stroke-induced muscle issues, impairs speech control. Stroke is a widely recognized public health issue, marked by a substantial occurrence and fatality rate \cite{feigin2022world}. Dysarthria affects half of stroke patients, making communication difficult \cite{flowers2013incidence}. Symptoms vary by stroke location and size, necessitating personalized therapy \cite{comrie2001influence,mitchell2018feasibility}. Current assessment relies on time-consuming auditory evaluation by healthcare professionals, highlighting the need for efficient automatic methods to improve stroke patient speech-language therapy \cite{perceptual_evaluation}.
We explore dysarthria assessment using the \textit{Autumn paragraph} in Korea \cite{kim2005dysarthria}. We focus on identifying Inappropriate Pauses (IPs), which are essential for patient feedback, highlighting areas requiring further training. Traditionally, speech-language pathologists handle this in their face-to-face sessions, but we propose automating pause detection and appropriateness assessment with an artificial intelligence model to support their work.

Inappropriate pause (IP) refers to delays that occur in untypical locations. In dysarthric speech, pauses come at unexpected locations, such as in the middle of a noun phrase, resulting in reduced speech intelligibility \cite{han2008breathgroup, inap-pause-labelling}. 
During the reading task, it is observed whether the patient cannot read a word in a single breath or stutters too much. Traditional assessment by pathologists relies on auditory analysis, lacking automatic inappropriate pause detection research. Detecting and assessing pauses within sentences using speech signals is necessary to evaluate pause appropriateness automatically. This is because the location of a pause within a sentence determines its appropriateness. Therefore, we first use the pause detection method to assess IPs. 

Some traditional dysarthric speech pause detection approaches \cite{acoustic-pause-1, Green2004-en} rely on algorithmic methods using amplitude thresholds. While these approaches effectively detect pause intervals, they only provide temporal information about pauses, not the location of pauses within the sentence. Some researchers use forced alignment to get pause locations in speech signals \cite{yeung15_interspeech,Diwakar2020}. Forced alignment detects pauses using speech and its transcription. However, its performance worsens because of the characteristics of dysarthric speech (i.e., inaccurate articulations, disfluencies, slow speaking rate, etc.) \cite{yeung15_interspeech,Diwakar2020}. Moreover, accurate transcription is necessary for forced alignment, resulting in a two-stage process: obtainment transcription and forced alignment. 

To tackle these challenges, we approach pause detection as a speech recognition problem. We treat pauses as a distinct token in the automatic speech recognition (ASR) model, which inputs speech and produces text with pause tags. 
Additionally, we enhance the ASR model with an inappropriate pause prediction layer, creating an end-to-end solution for detecting inappropriate pauses in dysarthric speech.
Our approach simplifies pause labeling at the text level compared to previous conventional temporal labeling. 
Furthermore, in collaboration with speech-language pathologists, we establish criteria for inappropriate pauses, focusing on within-word insertions for precise sentence pause identification.  
We also introduce a tailored evaluation metric specific to this task, allowing separate assessment of pause detection performance from ASR performance. From our experiment, incorporating pause detection into the ASR model also improves ASR performance.

\section{Labelling Pause Location and Inappropriate Pause in Dysarthric Speech}

We utilize the Korean dysarthric speech corpus, which comprises 2,251 utterances from the \textit{Autumn paragraph}, recorded by stroke patients. \textit{Autumn paragraph} contains all the necessary consonants and vowels for evaluation. Although recorded speeches are for paragraph reading, we experiment on the sentence level. This is because pauses within sentences carry more importance than those between sentences. The severity scale for this datset is the NIH Stoke Scale: without dysarthria, mild to moderate, and severe cases. Table \ref{tab:bi-corpus-stats} shows Korean dysarthric speech corpus statistics. We first label pause location at a text level to detect IP in dysarthric speech and then annotate each pause appropriateness. 

\begin{table}[th]
\caption{Dataset statistics by severity of dysarthric speech corpus.}
\centering
\resizebox{\columnwidth}{!}{%
\begin{tabular}{l|c|c|c|l}
\hline
Severity & w/o dysarthria & Mild-to-Moderate & Severe & Total \\ \hline \hline
\# of Utterances & 72 & 1985 & 194 & 2251 \\ \hline
\end{tabular}%
}
\label{tab:bi-corpus-stats}
\end{table}

\subsection{Labeling Pause Location}

\begin{figure}[t]
    \centering
    \includegraphics[width=\columnwidth]{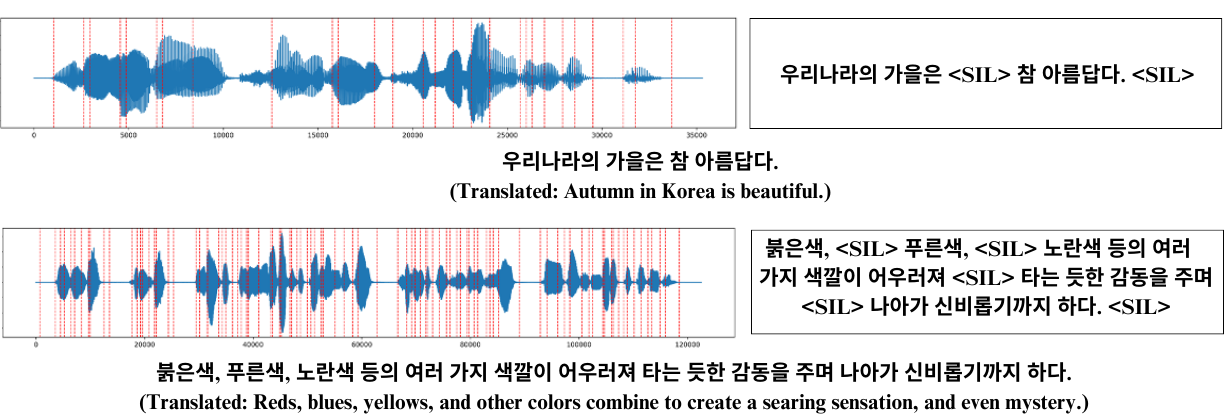}
    \caption{Examples of TIMIT-style labeling (left) and the proposed labeling (right). The samples are from the Korean dysarthric speech corpus.}
\label{fig:labelling-vs}
\end{figure}

Following previous research \cite{inap-pause-labelling, pause-definition, pause-definition-2}, we define a pause as an uninterrupted silence portion of an utterance lasting at least 150 msec. We add \texttt{<SIL>} tags to indicate pause locations in the text level for the pause detection task, as shown in Figure \ref{fig:labelling-vs}. In Figure \ref{fig:labelling-vs}, we compare the proposed method to the TIMIT \cite{garofolo1993timit}-style method, which labels the time information of each phoneme and pauses. TIMIT-style labeled corpus is usually utilized in phoneme segmentation. This approach is costly in labeling, but a substantial amount of labeled corpus is required to achieve reliable performance for clinical purposes. Therefore, we propose a simplified labeling method that reduces costs and enables more efficient analysis of the position of pauses on the paragraph-reading task with a smaller labeled corpus. 

With forced alignment, the pause location labeling is not necessary. However, forced alignment requires a transcription of a speech. In other words, forced alignment still needs transcription labeling. Our labeling strategy, which annotates pause location at the text level, allows for pause location labeling at a cost similar to transcription labeling.

\begin{figure*}[ht]
    \centering
    \includegraphics[width=1.0\textwidth]{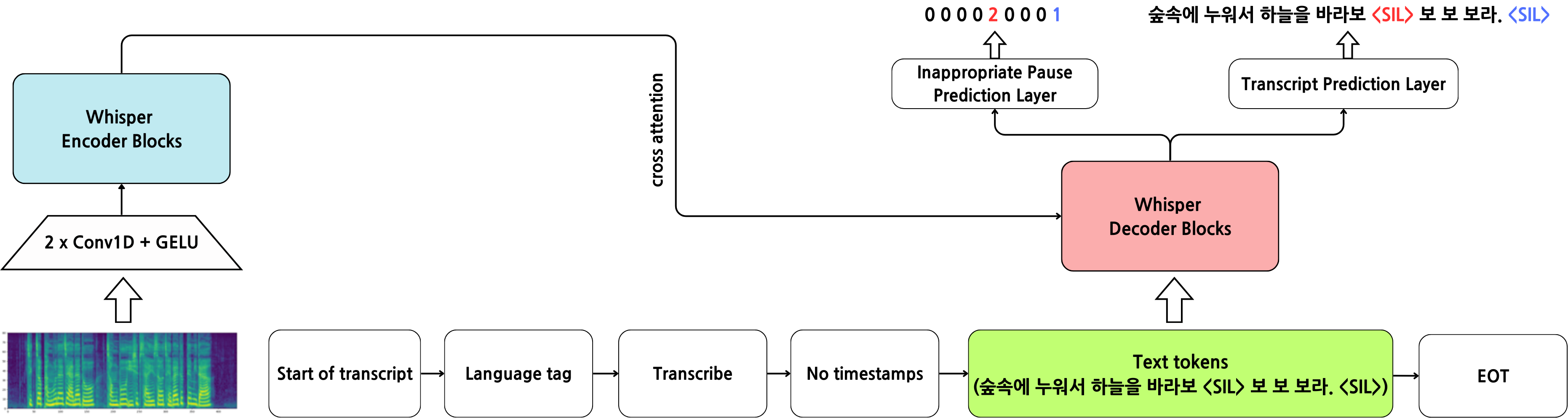}
    \caption{The proposed inappropriate pause detection model architecture. Above the whisper decoder layers, there are two task-specific layers: The inappropriate Pause Prediction layer (IP Prediction layer) and the Transcript Prediction Layer.}
\label{fig:model_architecture}
\end{figure*}

\subsection{Labeling Inappropriate Pauses in Dysarthric Speech}
\label{labelling inappropriate pauses}

Based on the annotated locations of pauses, we further label their appropriateness. The labeling criteria were established in collaboration with speech-language pathologists based on their professional expertise in speech therapy. IP was annotated at the word level, where non-pauses are marked as 0, appropriate pauses as 1, and inappropriate pauses as 2. (\textit{IPSeq} in Figure \ref{fig:bi_sequence}.)

We define inappropriate pauses into three categories: Inappropriate pauses within a word unit, Pauses following vocalic surplus expressions, and Pauses resulting from attempts to correct inaccurate pronunciation. First, Inappropriate pauses within a word unit are inappropriate pauses. If pauses occur in unexpected places, such as within a noun phrase, they can detract from readability and intelligibility. Therefore, the pause is considered inappropriate if a patient cannot say a word in a single breath. The IPs corresponding to the first category pauses that occur at unrelated syntax boundaries are our primary concern \cite{inap-pause-labelling, ip-syntactic}. Second, pauses following vocalic surplus expressions such as “uh” or “um” are inappropriate. Vocalic surplus expressions due to hesitation are not the focus of speech therapy, so only pauses accompanying unclear wording or incorrect pronunciation and excessively long pauses are deemed inappropriate. The excessively long pauses are pauses longer than 3 seconds. Lastly, pauses that occur to rectify mispronunciation are marked as inappropriate rather than simple repetitions. Patients tend to repeat the syllables because they are hard to pronounce words correctly at once. This criterion was also applied only when the pause intervals were excessively long. (longer than 3 seconds)

\section{Inappropriate Pause Detection as ASR}

We define pause detection as a speech recognition task and employ sequence-to-sequence (Seq2Seq) architecture to generate transformed text (text with pause tags) for a given speech, as shown in Figure \ref{fig:model_architecture}. We use OpenAI's whisper-small \cite{radford2022robust} as our speech Seq2Seq model. Furthermore, we extend the Whisper model to train inappropriate pause detection and speech recognition tasks simultaneously. The detailed process of the proposed method is shown in Equation \ref{eq:projection}. First, the Seq2Seq model processes an input speech X. From the latent representation Z of the Seq2Seq model, we incorporate two separate layers: the projection layer to its vocabulary size (Transcript prediction layer) and the inappropriate pause prediction layer (IP prediction layer). The transcript prediction layer projects the last hidden state to the vocabulary size for input speech transcription. Its output V includes pause tags if there is a pause interval in X. The IP prediction layer determines whether each token is an appropriate pause, inappropriate pause, or non-pause (word). To train the IP prediction layer, we employ the soft-dtw loss. The final loss $L$ is shown in Equation \ref{eq:loss}. Our model detects pauses and determines their appropriateness in an end-to-end manner, eliminating the need for post-processing.

\begin{equation}
\resizebox{0.5\linewidth}{!}{$
    \begin{aligned}
    Z &= \text{Seq2Seq}(X) \\
    V' &= \text{TranscriptPrediction}(Z) \\
    IP' &= \text{IPPrediction}(Z)
    \end{aligned}
$}
\label{eq:projection}
\end{equation}

\begin{equation}
L = \text{Cross-entropy}(V') + \text{Soft-dtw}(IP')
\label{eq:loss}
\end{equation}

\section{Experiments}

\subsection{Experimental Setup}
\label{exp:set-up}

The corpus used for training is divided into training, validation, and test sets in the ratio of 8:1:1. The samples for each set are 1800, 225, and 226, respectively. The division of the dataset is done while maintaining the ratio of data samples for each severity level.
The hyper-parameters used for training are as follows. For the AdamW \cite{adamw} optimizer, we set the learning rate to 5e-4 with linear decaying. We set the batch size as 8. 

We compare the proposed method with Montreal Forced Alignment (MFA)\cite{MFA}.
For baseline, three types of MFA were used. \textit{MFA-GT} aligns the ground truth transcription and speech. \textit{MFA-Whisper} and \textit{MFA-Dysarthric-Whisper} utilize ASR transcription from open-source Whisper and fine-tuned Whisper with dysarthric speech, respectively. (Table \ref{tab:result}) Whisper fine-tuning was done with the same Korean dysarthric speech corpus, and the training details were the same. 
Moreover, we also train the proposed method, except with a different ASR model, Conformer-RNNT \cite{gulati20_interspeech}. We used a Korean pre-trained Conformer model.\footnote{\url{https://huggingface.co/eesungkim/stt_kr_conformer_transducer_large}}
The training details were maintained for the Conformer-RNNT model. Note that \textit{Conformer-RNNT} does not have an IP prediction layer. We further discuss this in Section \ref{sec:discuss}. Note that we only compare the pause detection performance with other methods because, to our best knowledge, we are the first to detect IPs in dysarthric speech automatically. Hence, there is no baseline.
 
\begin{figure}[]
\centering
 \includegraphics[width=\columnwidth]{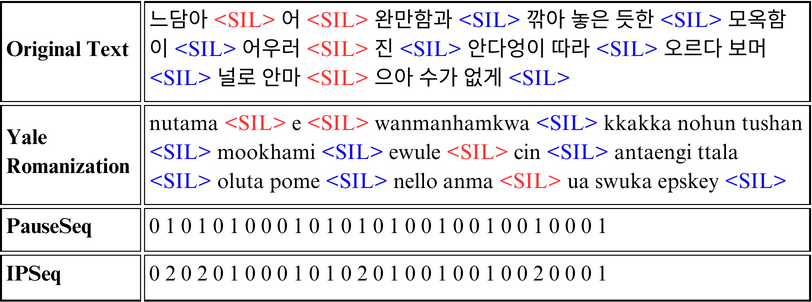}
 \caption{Example of pause sequences for calculation. Above is an example of a sequence with pauses to see if the model measures pauses well, and below is an example of a sequence to see if the model measures inappropriate pauses well.}
\label{fig:bi_sequence}
\end{figure}

\subsection{Evaluation Metrics}

We separately evaluate ASR and pause detection. We use word error rate (WER) and character error rate (CER) for ASR. To measure WER and CER, we remove pause tags from a generated transcript. In the case of pause detection, we propose new metrics to evaluate the accuracy of pause detection and inappropriate pause detection tasks independently of the speech recognition results. To this end, we first construct \textit{PauseSeq} and \textit{IPSeq} as shown in Figure \ref{fig:bi_sequence}. In the case of \textit{PauseSeq}, the ground truth and predicted transcriptions are marked as 1 if a word is a pause tag and 0 otherwise. For \textit{IPSeq}, which stands for Inappropriate Pause Sequence, each transcription is converted to the sequence with 0, 1, and 2, where 0 is non-pause, 1 is an appropriate pause, and 2 is IP. From the asr-independent sequence, we measure the performance by error rate. We calculate CER with the \textit{PauseSeq} and \textit{IPSeq} and refer to these \textit{PauER} and \textit{IPER}, respectively.

\subsection{Results}
\label{exp:pause-detection}


Table \ref{tab:result} shows the ASR and pause detection performance. Our method performs better than other methods in ASR and pause detection. Specifically, \textit{PauER} is lower in our method than \textit{MFA-GT}. When comparing three types of MFA, pause detection performance was influenced by transcription performance. The best results among MFAs were obtained using GT transcription, followed by \textit{Dysarthric-Whisper} and \textit{Whisper}. This trend persisted even for pause detection fine-tuned \textit{Conformer-RNNT} model. Higher ASR performance correlated with improved pause detection performance. Additionally, comparing ASR performance, fine-tuning with dysarthric speech for ASR-only (\textit{MFA-Dysarthric-Whisper}) versus jointly training for pause detection and ASR (\textit{Ours}) showed enhanced ASR performance. This suggests a complementary relationship between pause detection and ASR in dysarthric speech, where prolonged pause intervals are prominent characteristics.

\begin{table}[]
\centering
\caption{Pause detection and ASR evaluation. \textit{WER} and \textit{CER} measure ASR performance, and \textit{PauER} measures pause detection performance.}
\label{tab:result}
\resizebox{\columnwidth}{!}{%
\begin{tabular}{l|c|c|c}
\hline
                       & WER(\%)        & CER(\%)        & PauER(\%)      \\ \hline \hline
MFA-GT                 & -              & -              & 11.14          \\ \hline
MFA-Whisper            & 54.89          & 27.35          & 22.49          \\ \hline
MFA-Dysarthric-Whisper & 32.21          & 22.38          & 17.27          \\ \hline \hline
Conformer-RNNT         & 64.52          & 49.99          & 22.81          \\ \hline
Ours                   & \textbf{25.31} & \textbf{11.96} & \textbf{3.077} \\ \hline
\end{tabular}%
}
\end{table}

\begin{table}[t]
\centering
\caption{IP detection, Pause detection, and ASR performance according to severity level}
\label{tab:sev-perf}
\resizebox{\columnwidth}{!}{%
\begin{tabular}{l|c|c|c|c}
\hline
Severity         & WER(\%) & CER(\%) & PauER(\%) & IPER(\%) \\ \hline \hline
Total            & 25.31   & 11.96   & 3.07      & 14.47    \\ \hline \hline
w/o dysarthria   & 6.93    & 2.89    & 2.48      & 20.69    \\ \hline
Mild-to-Moderate & 22.38   & 10.20   & 3.03      & 15.53    \\ \hline
Severe           & 57.44   & 30.47   & 3.60      & 13.40    \\ \hline
\end{tabular}%
}
\end{table}

Table \ref{tab:sev-perf} shows the IP detection performance of the proposed method. \textit{Total} is the same model with \textit{Ours} in Table \ref{tab:result}. We dissect the performance according to the dysarthria severity. Dysarthric speech becomes more slurred and challenging to understand as its severity increases. An effective IP detection method operates robustly across different levels of severity. This is crucial to ensure that effective diagnosis and feedback can be provided even for patients with severe dysarthria. In Table \ref{tab:sev-perf}, we can observe that IP and pause detection performance remains relatively consistent across different severity levels. In the case of \textit{IPER}, performance deteriorates as severity decreases. This is because lower severity levels result in fewer instances of IPs. However, ASR performance decreases as the severity level increases. This suggests that the data used for training was sufficient for IP detection but insufficient for training ASR, leading to this discrepancy in performance.

\section{Discussion}
\label{sec:discuss}
We experimented with our proposed architecture across various ASR models, including wav2vec 2.0 \cite{NEURIPS2020_92d1e1eb}-CTC, but the IP layer was not trained effectively with decoding methods such as CTC and RNN-Transducer. It's possible that our proposed IP detection method may not apply to ASR models with different decoding strategies. However, the approach using Whisper, as we suggest, offers scalability. With simple text-level labeling, it can be extended to languages beyond Korean. Moreover, incorporating recent advancements in Whisper, such as word-boundary extraction using cross-attention, it becomes possible to extract pause durations as well.

\section{Conclusion}
We address IP detection in dysarthric speech by treating pauses as distinct tokens in the ASR model. Furthermore, we extend the ASR model with the IP prediction layer, constructing an end-to-end process in IP detection. In this way, we simplify pause labeling and enhance ASR performance in dysarthric speech. We also establish pause appropriateness criteria in the collaboration of medical professionals.
The detailed evaluation showed that the proposed method performs better pause detection than baseline and severity-robust IP detection performance. The proposed model's consistency in detecting inappropriate pauses across different dysarthria severity levels highlights its potential for providing effective diagnosis and feedback, particularly for patients with severe dysarthria.

\bibliographystyle{IEEEbib}
\bibliography{refs}

\end{document}